\documentclass[conference]{IEEEtran}
\IEEEoverridecommandlockouts
\pdfoutput=1
\usepackage{cite}
\usepackage{amsmath,amssymb,amsfonts}
\usepackage{algorithmic}
\usepackage{graphicx}
\usepackage{textcomp}
\usepackage{xcolor}

\usepackage{bm}
\usepackage{comment}
\usepackage{multirow}

\usepackage{pgfplots}

\pgfplotsset{compat=1.18}
\usepackage{subcaption}

\usepackage{fancyhdr}

\DeclareMathOperator{\sinc}{sinc}

\def\BibTeX{{\rm B\kern-.05em{\sc i\kern-.025em b}\kern-.08em
    T\kern-.1667em\lower.7ex\hbox{E}\kern-.125emX}}
\begin{document}

\title{Realistic Scatterer Based Adversarial Attacks on SAR Image Classifiers\\
\thanks{
Accepted by the IEEE International Radar Conference 2023.

© 2023 IEEE. Personal use of this material is permitted. Permission from IEEE must be
obtained for all other uses, in any current or future media, including
reprinting/republishing this material for advertising or promotional purposes, creating new
collective works, for resale or redistribution to servers or lists, or reuse of any copyrighted
component of this work in other works.}
}

\author{
\IEEEauthorblockN{Tian Ye\IEEEauthorrefmark{1}, Rajgopal Kannan\IEEEauthorrefmark{2}, Viktor Prasanna\IEEEauthorrefmark{1}, Carl Busart\IEEEauthorrefmark{2}, Lance Kaplan\IEEEauthorrefmark{2}}
\IEEEauthorblockA{
    \IEEEauthorrefmark{1}University of Southern California \IEEEauthorrefmark{2}DEVCOM Army Research Lab\\
    \IEEEauthorrefmark{1}\{tye69227, prasanna\}@usc.edu \IEEEauthorrefmark{2}\{rajgopal.kannan.civ, carl.e.busart.civ, lance.m.kaplan.civ\}@army.mil}
}

\maketitle

\begin{abstract}
Adversarial attacks have highlighted the vulnerability of classifiers based on machine learning for Synthetic Aperture Radar (SAR) Automatic Target Recognition (ATR) tasks. An adversarial attack perturbs SAR images of on-ground targets such that the classifiers are misled into making incorrect predictions. However, many existing attacking techniques rely on arbitrary manipulation of SAR images while  overlooking the feasibility of executing the attacks on real-world SAR imagery. Instead, adversarial attacks should be able to be implemented by physical actions, for example, placing additional false objects as scatterers around the on-ground target to perturb the SAR image and fool the SAR ATR.

In this paper, we propose the On-Target Scatterer Attack (OTSA), a scatterer-based physical adversarial attack. To ensure the feasibility of its physical execution, we enforce a constraint on the positioning of the scatterers. Specifically, we restrict the scatterers to be placed only on the target instead of in the shadow regions or the background. To achieve this, we introduce a positioning score based on Gaussian kernels and formulate an optimization problem for our OTSA attack. Using a gradient ascent method to solve the optimization problem, the OTSA can generate a vector of parameters describing the positions, shapes, sizes and amplitudes of the scatterers to guide the physical execution of the attack that will mislead SAR image classifiers.
The experimental results show that our attack obtains significantly higher success rates under the positioning constraint compared with the existing method.

\end{abstract}


\section{Introduction}
\label{sec:intro}


Synthetic Aperture Radar (SAR) Automatic Target Recognition (ATR) is a critical technique in remote sensing image recognition that finds widespread application in real-world scenarios~\cite{sar1,sar2,sar3}. SAR images are generated by processing reflected radar signals from sensors and are passed on to ATR systems for real-time object recognition. Efficient and accurate ATR of SAR images is essential for enhanced situational awareness, particularly in military applications. 
Machine learning techniques have been successfully used for SAR image classification tasks, with Convolutional Neural Networks (CNNs)~\cite{cnn1,cnn3,cnn4} being the most popular due to their strong feature extraction capabilities. More recently, Graph Neural Networks (GNNs)~\cite{gnn} have also been proved to be effective as classifiers for SAR images.
These deep learning models can learn highly discriminative features from large-scale SAR datasets, enabling accurate and efficient ATR of SAR images.


Recent research~\cite{sar-attack1,sar-attack2,sar-attack3,sar-attack4,sar-attack5} has shown that SAR ATR is vulnerable to adversarial attacks. Adversarial attacks on SAR ATR involve adding carefully crafted noise to perturb the input SAR image, which is similar to the original one but will cause the classifiers to misclassify the image. In military applications, adversaries can exploit these attacks to deceive image classifiers of SAR ATR, leading to incorrect target recognition and potentially causing serious consequences. 
On the other hand, adversarial attacks can be useful for generating perturbed datasets that can be used to train the classifiers for SAR ATR, ultimately improving their robustness to attacks from potential adversaries.

Feasibility is a critical challenge for adversarial attacks on SAR image classifiers. Most existing attacks focus on perturbing SAR images without considering how the manipulations can be practically implemented. As SAR images are generated by the SAR system internally, it is hard for adversaries to break into the system and arbitrarily manipulate the SAR images. 
Instead, it is more practical for the adversary of SAR ATR to physically execute the attack by, for example, attaching ``false" objects near the on-ground target. The false objects act as scatterers that will be captured into the SAR images and make perturbation on the original image. If the scatterers are carefully designed, it is possible to mislead the classifier to make wrong predictions. Attacks of this nature are called \textit{physical adversarial attacks}. 
The key point of this type of attacks is to design scatterers that will effectively fool the SAR image classifier.
Specifically, their goal is to generate an ``authentic" SAR image through pixel manipulation that corresponds to the scattered image obtained by the presence of false objects near the physical target. They should simulate the perturbed SAR image as if the false objects (as scatterers) are attached to the on-ground target or the background around the target, and evaluate if the perturbed SAR image will mislead the classifier to make a wrong prediction.

\begin{figure}
  \centering
  \includegraphics[width=0.4\textwidth]{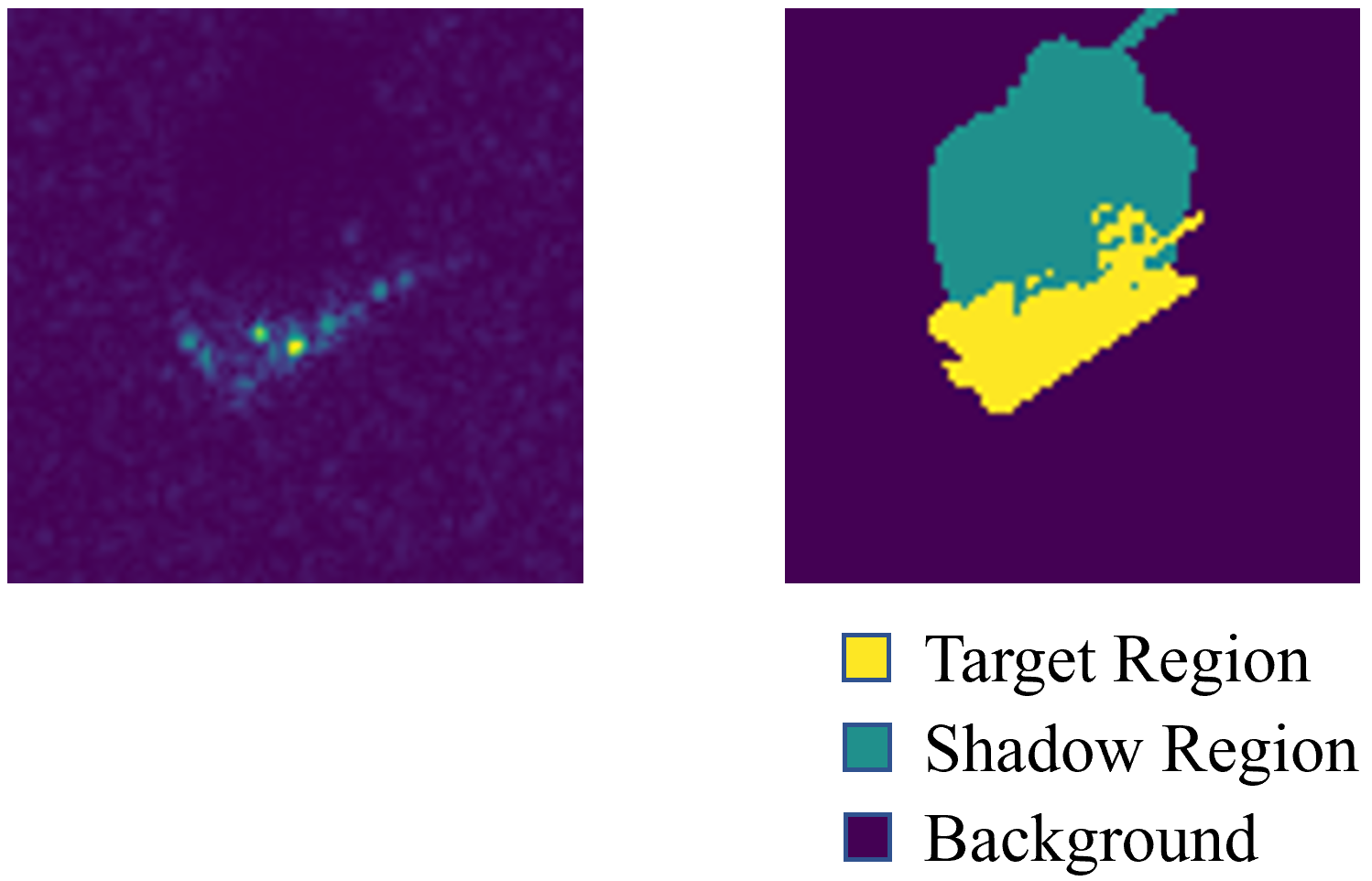}
  \caption{Left: A SAR image from the MSTAR~\cite{mstar} dataset. Right: The target region, shadow region and background of the SAR image.}
  \label{fig:region}
\end{figure}

In this paper, we propose a physical adversarial attack called \textbf{On-Target Scatter Attack (OTSA)}. We enforce a positioning constraint on the scatterers to ensure its feasibility. Specifically, all scatterers are required to be in the region of the on-ground target instead of the shadow region or the background, as illustrated in Figure~\ref{fig:region}. This constraint offers two advantages: First, the scatterers can cause stronger perturbation on the SAR images when they are on the on-ground target, rather than in the shadow regions. Second, it enables future research on adversarial attacks on SAR images of moving targets as the scatterers will be able to move along with the on-ground target. Our main contributions are summarized as follows:
\begin{enumerate}
    \item We define the threat model of scatterer-based physical adversarial attacks on SAR image classifiers. We further introduce a constraint on the positioning of the scatterers to ensure the feasibility of the attacks. We discuss the motivation and the importance of the constraint that allows more feasible physical execution of adversarial attacks on SAR image classifiers. 
    \item We propose the On-Target Scatterer Attack (OTSA) with positioning scores based on Gaussian kernels and formulate it into a maximization problem that can be solved by the gradient ascent algorithm.
    \item We implement our OTSA attack and conduct experiments over the MSTAR dataset to attack nine different classifiers for SAR images. Experimental results show significant improvements in terms of the success rate in our OTSA attack compared with the baseline.
\end{enumerate}

\section{Problem Definition and Preliminary}
\subsection{Problem Definition and Threat Model}
\label{sec:threat}

In the scenario of Synthetic Aperture Rader (SAR) Automatic Target Recognition (ATR), the SAR system sends radar signals from the air to the on-ground targets, collects the reflections and generates the SAR images of the targets. We assume there is a classifier denoted by $F(\cdot)$ inside the SAR system to recognize the targets in the SAR images. For an input SAR image denoted by $\mathbf{X}$ and its ground truth class denoted by $y$, presumably the classifier can correctly predict the class of $\mathbf{X}$, i.e., $F(\mathbf{X})=y$.

An \textit{adversary} of the SAR system is the party who aims at protecting the on-ground target from being recognized by the SAR system by attacking the classifier $F(\cdot)$.
For an on-ground target, the goal for the adversary is to perturb the SAR image $\mathbf{X}$ into $\mathbf{X}^{\text{adv}}$ such that the classifier will be fooled to give a wrong prediction, i.e., $F(\mathbf{X}^{\text{adv}})\ne y$.

Although traditional pixel-wise manipulations~\cite{fgsm}\cite{i-fgsm} have been shown effective to fool classifiers for natural images, they are not feasible for SAR ATR because it is hard for the adversary to break into the SAR system and arbitrarily manipulate the SAR images. Instead, the adversary should perturb the SAR image by placing a few scatterers as ``false" objects near the on-ground target. We call this method a \textit{scatterer-based physical adversarial attack}. To provide a guidance for the adversary, our problem setting focuses on figuring out how the scatterers should be placed in terms of their shapes, sizes, positions, amplitudes and orientations. 

To this end, we use a scattering model~\cite{scatter-attack} (described in Section~\ref{sec:scattering}) to simulate the SAR image of any given scatterers. We propose an adversarial attack that generates the SAR image through pixel manipulation that corresponds to the scattered image obtained by the presence of false objects near the target. Our method will be a guidance for the adversary to physically place the scatterers near the target to feasibly apply the attack.

\renewcommand{\arraystretch}{1.3}

\begin{table}[htbp]
\caption{Parameters in the Scattering Model}
\begin{center}
\begin{tabular}{|c|c|c|}
\hline
 & \textbf{Parameter} & \textbf{Explanation} \\
\hline
\multirow{7}{6em}{Parameters $\bm{\theta}$ for a scatterer}&$A$ & Amplitude \\
\cline{2-3}
&$x$ & Range position \\
\cline{2-3}
&$y$ & Cross-range position \\
\cline{2-3}
&$\alpha$ & Frequency dependence \\
\cline{2-3}
&$L$ & Length \\
\cline{2-3}
&$\bar{\phi}$ & Orientation \\
\cline{2-3}
&$\gamma$ & Aspect dependence \\
\hline
\multirow{10}{6em}{Parameters $\bm{\xi}$ for SAR imaging}&$B$ & Bandwidth of radar wave\\
\cline{2-3}
&$f_c$ & Center frequency of radar wave\\
\cline{2-3}
&$c$ & Velocity of light\\
\cline{2-3}
&$\phi_m$ & Aperture accumulation angle\\
\cline{2-3}
&$\eta_x$ & Factor $(m-1)/(m^*-1)$\\
\cline{2-3}
&$\eta_y$ & Factor $(n-1)/(n^*-1)$\\
\cline{2-3}
& \multirow{2}{*}{$p_x$} & Pixel spacing of range \\
&& $c\cdot\eta_x/(2B)$\\
\cline{2-3}
& \multirow{2}{*}{$p_y$} & Pixel spacing of cross-range \\
&&$c\cdot\eta_y/(4f_c\sin(\phi_m/2))$\\
\hline
\end{tabular}
\label{tab:param}
\end{center}
\end{table}

\subsection{Scattering Model}
\label{sec:scattering}
The Attributed Scattering Center Model (ASCM)~\cite{ascm,scatter-attack} estimates how a scatterer on the ground will appear in a SAR image. An scatterer is characterized with a vector of seven parameters $\bm{\theta}=[A, x, y, \gamma, L, \alpha, \bar{\phi}]$, which is summarized in Table~\ref{tab:param}. The parameters describe the scatterer in terms of its amplitude, position, shape, length, orientation, etc. Given a set of $N$ scatterers parameterized by $\bm{\Theta}_N=\{\bm{\theta}_i=[A_i,x_i,y_i,\gamma_i,L_i,\alpha_i,\bar{\phi}_i]\mid i=1,...,N\}$, the ASCM converts them into SAR images in the following steps:

1) The total backscattered field from all the scatterers denoted by $E_{m\times n}(f, \phi; \bm{\Theta}_N)$ are computed as in terms of frequency $f$ and aspect angle $\phi$, where $m$ and $n$ are the number of samples along the dimension of $f$ and $\phi$ respectively.

2) The backscattered field is then transferred from the polar plane to the Cartesian plane by uniformly resampling as $E_{m^{\ast}\times n^{*}}(f_x,f_y;\bm{\Theta}_N)$, where $f_x=f\cos(\phi)$ and $f_y=f\sin(\phi)$. $m^*$ and $n^*$ are the number of samples along the dimension of $f_x$ and $f_y$ respectively. The equation of $E_{m^*\times n^*}(f_x,f_y;\bm{\Theta}_N)$ is given by

\begin{equation}
    E_{m^*\times n^*}(f_x,f_y;\bm{\Theta}_N)
=\sum_{i=1}^NE_i(f_x,f_y;\bm{\theta}_i)
\end{equation}
where
     \begin{align}
     \label{eq:ascm}
        \begin{split}
&E_i(f_x,f_y;\bm{\theta}_i)\\
=& A_i\cdot\left(\frac{j\sqrt{f_x^2+f_y^2}}{f_c}\right)^{\alpha_i}\cdot\exp\left(-\frac{f_y}{f_c}\gamma_i\right)\\
\cdot&\sinc\left(\frac{\pi\sqrt{f_x^2+f_y^2}}{2\sin{(\phi_m/2)}}L_i\eta_y\sin{\left(\tan^{-1}\left(\frac{f_y}{f_x}\right)-\bar{\phi_i}\phi_m/2\right)}\right)\\
\cdot&\exp\left(-\frac{j4\pi}{c}(p_xx_if_x+p_yy_if_y)\right)
        \end{split}
    \end{align}
The parameters used in Equation (\ref{eq:ascm}) related to SAR imaging are summarized in Table~\ref{tab:param}.

3) A two-dimensional inverse discrete Fourier transformation (2D-IDFT) is applied to $E(f_x,f_y;\bm{\Theta}_N)$ which outputs a SAR image of the scatterers denoted by $I(\bm{\Theta}_N, \bm{\xi})$.

In summary, the entire process takes as input the parameters $\bm{\Theta}_N$ about the $N$ scatterers and the parameters $\bm{\xi}$ about the SAR imaging and generates a SAR image $I(\bm{\Theta}_N, \bm{\xi})$ for the scatterers as output.

\section{Proposed Method}
\label{sec:proposed}
\subsection{Positioning Constraint}
\label{sec:motivation}
As described in our threat model in Section~\ref{sec:threat}, our physical adversarial attack will simulate the SAR image of the on-ground target as if there are additional scatterers placed nearby to guide the adversary to physically manipulate the SAR image. In terms of the positioning of the scatterers, the shadow regions besides the target would be particularly favored by adversarial attacks as they contain the majority of the structural information of the target. 
However, for the sake of feasibility, we should enforce a constraint that the scatterers must be placed onto the target instead of the shadow regions. There are at least three motivations to introduce this constraint:
\begin{enumerate}
    \item It is hard to physically execute the attack if scatterers are in the shadow regions. The shadow region is the area where radar signals have low reflections available to the SAR aircraft. Scatterers placed in the shadow region will reflect low signals and thus appear hardly invisible in the SAR image.
    \item The position and shape of the shadow region is variant to the relative direction of the on-ground target with respect to the SAR aircraft. Therefore, constraining the scatterers on the target is necessary to develop feasible attacks invariant to the direction of the SAR aircraft.
    \item Scatterers attached and sticked to the target can be carried along with the movement of the target. 
\end{enumerate}
Though we currently consider SAR images taken from fixed directions and for static targets only, the constraint will be a step towards attacking SAR images for direction-invariant SAR and moving targets in the future.

\begin{table*}[htbp]
\caption{Classifiers Trained over the MSTAR Dataset}
\begin{center}
\begin{tabular}{|c|c|c|c|c|c|c|c|c|c|}
\hline
 & GNN~\cite{gnn} & VGG11 & AlexNet & SqueezeNet & DenseNet121 & MobileNetv2 & ResNet50 & ShuffleNetv2 & AConvNet\\
\hline
Accuracy (\%) & 94.4 & 98.2 & 94.7 & 98.6 & 97.6 & 98.6& 98.5 & 97.2 & 98.3 \\
\hline
\end{tabular}
\label{tab:classifiers}
\end{center}
\vspace{-0.3cm}
\end{table*}

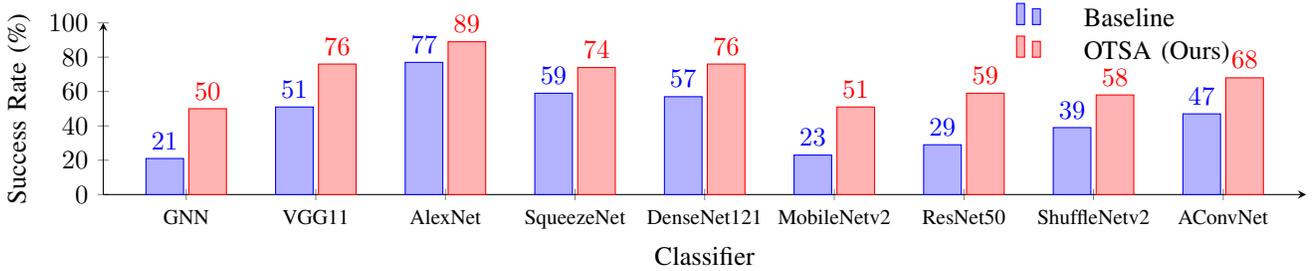
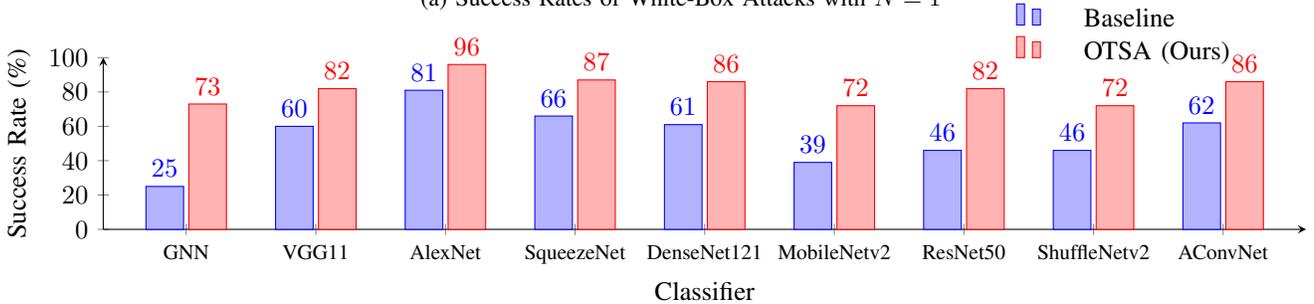
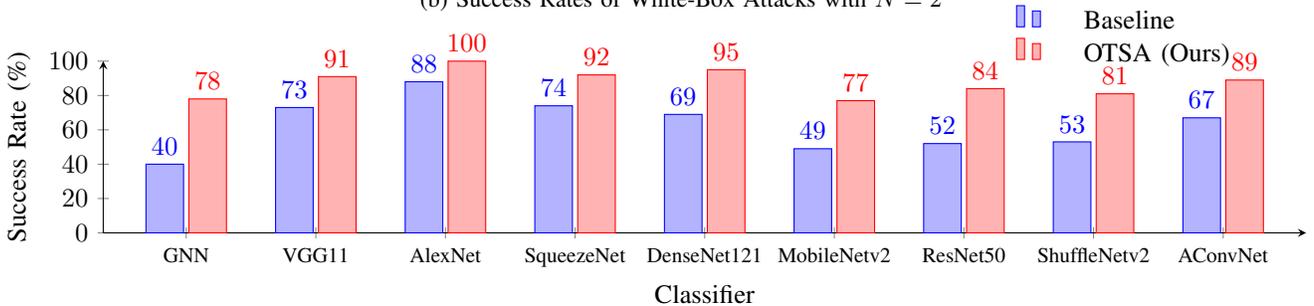
\begin{figure*}[htbp]
    \centering
    \begin{subfigure}[b]{\textwidth}
  \begin{tikzpicture}
    \begin{axis}[      
        ybar,      
        width=500pt,
        height=110pt,
        ylabel=Success Rate (\%),      
        xlabel=Classifier,      
        xticklabels={GNN,VGG11,AlexNet,SqueezeNet,DenseNet121,MobileNetv2,ResNet50,ShuffleNetv2,AConvNet},      
        xtick={1,2,3,4,5,6,7,8,9},   
        xticklabel style={font=\footnotesize},
        ymin=0,      
        ymax=100,      
        axis lines=left,
        bar width=0.5cm,      
        enlarge x limits=0.08,      
        legend style={at={(0.85,1.15)},     
        cells={anchor=west},
        anchor=north,legend columns=1,
        draw=none,
        fill=none,
        column sep=0.5cm},
        nodes near coords, 
        nodes near coords align={vertical},
    ]
      \addplot coordinates {(1,21) (2,51) (3,77) (4,59) (5,57) (6,23) (7,29) (8,39) (9,47)};
      \addplot coordinates {(1,50) (2,76) (3,89) (4,74) (5,76) (6,51) (7,59) (8,58) (9,68)};
      \legend{Baseline, OTSA (Ours)}
    \end{axis}
  \end{tikzpicture}
        \caption{Success Rates of White-Box Attacks with $N=1$}
        \label{fig:1-scatterer}
    \end{subfigure}
    
    \vspace{-0.2cm} 
    
    \begin{subfigure}[b]{\textwidth}
  \begin{tikzpicture}
    \begin{axis}[      
        ybar,      
        width=500pt,
        height=110pt,
        ylabel=Success Rate (\%),      
        xlabel=Classifier,      
        xticklabels={GNN,VGG11,AlexNet,SqueezeNet,DenseNet121,MobileNetv2,ResNet50,ShuffleNetv2,AConvNet},      
        xtick={1,2,3,4,5,6,7,8,9},   
        xticklabel style={font=\footnotesize},
        ymin=0,      
        ymax=100,      
        bar width=0.5cm,    
        axis lines=left,
        enlarge x limits=0.08,      
        legend style={at={(0.85,1.35)}, 
        cells={anchor=west},
        anchor=north,legend columns=1,
        draw=none,
        fill=none,
        column sep=0.5cm},
        nodes near coords, 
        nodes near coords align={vertical},
    ]
      \addplot coordinates {(1,25) (2,60) (3,81) (4,66) (5,61) (6,39) (7,46) (8,46) (9,62)};
      \addplot coordinates {(1,73) (2,82) (3,96) (4,87) (5,86) (6,72) (7,82) (8,72) (9,86)};
      \legend{Baseline, OTSA (Ours)}
    \end{axis}
  \end{tikzpicture}
        \caption{Success Rates of White-Box Attacks with $N=2$}
        \label{fig:2-scatterers}
    \end{subfigure}

        \vspace{-0.2cm} 
    
    \begin{subfigure}[b]{\textwidth}
  \begin{tikzpicture}
    \begin{axis}[      
        ybar,      
        width=500pt,
        height=110pt,
        ylabel=Success Rate (\%),      
        xlabel=Classifier,      
        xticklabels={GNN,VGG11,AlexNet,SqueezeNet,DenseNet121,MobileNetv2,ResNet50,ShuffleNetv2,AConvNet},      
        xtick={1,2,3,4,5,6,7,8,9},   
        xticklabel style={font=\footnotesize},
        ymin=0,      
        ymax=100,      
        bar width=0.5cm,     
        axis lines=left,
        enlarge x limits=0.08,      
        legend style={at={(0.85,1.36)},   
        cells={anchor=west},
        fill=none,
        anchor=north,legend columns=1,
        draw=none,
        column sep=0.5cm},
        nodes near coords, 
        nodes near coords align={vertical},
    ]
      \addplot coordinates {(1,40) (2,73) (3,88) (4,74) (5,69) (6,49) (7,52) (8,53) (9,67)};
      \addplot coordinates {(1,78) (2,91) (3,100) (4,92) (5,95) (6,77) (7,84) (8,81) (9,89)};
      \legend{Baseline, OTSA (Ours)}
    \end{axis}
  \end{tikzpicture}
        \caption{Success Rates of White-Box Attacks with $N=3$}
        \label{fig:3-scatterers}
    \end{subfigure}
    \caption{Performance of our OTSA attack and the baseline. The number of scatterers $N$ is set to 1, 2 and 3 respectively. The horizontal axis is for the nine different classifiers that are attacked. The vertical axis is the success rate of the attacks.}
    \label{fig:comparison}
    \vspace{-0.6cm}
\end{figure*}

\begin{figure*}[ht]
  \centering
  \includegraphics[width=\textwidth]{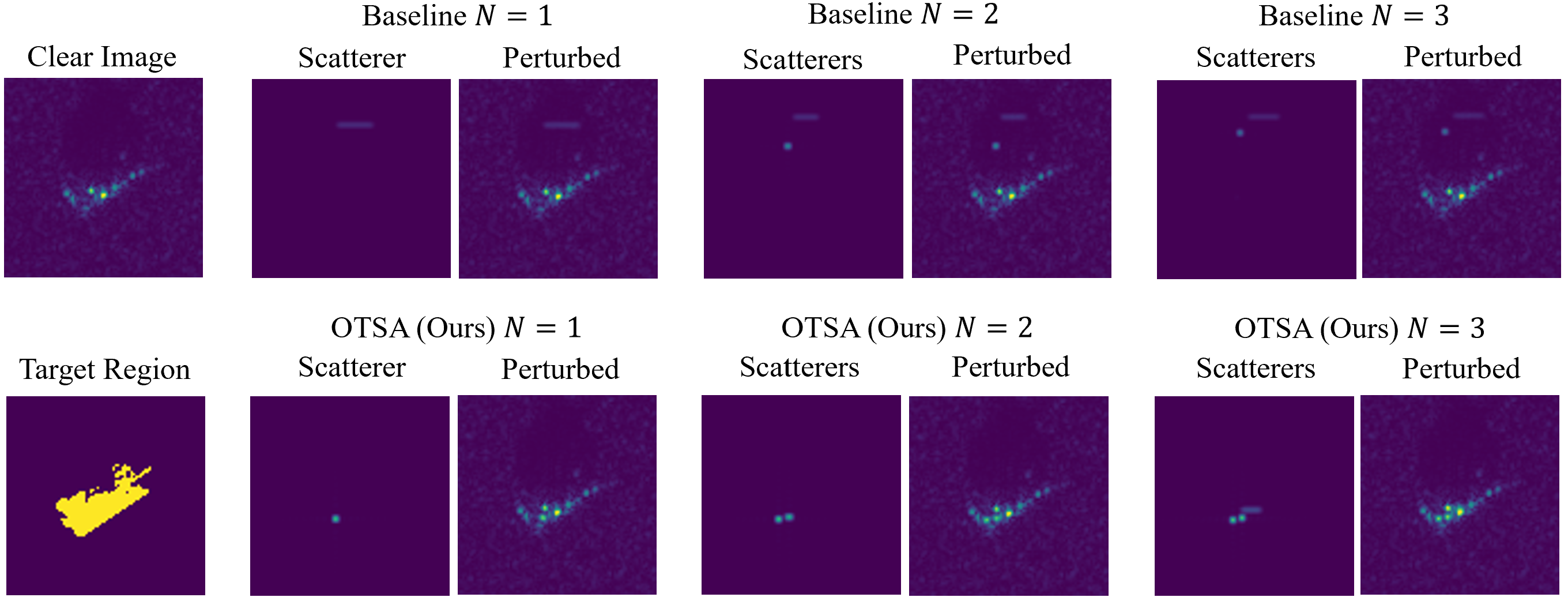}
  \caption{An example of SAR image attacked by our OTSA and the baseline using the GNN~\cite{gnn} classifier. ``Clear Image" is the original SAR image $\mathbf{X}$ to be attacked. ``Target Region" is a bitmap image for $\mathcal{M}^{\mathbf{X}}$ with yellow pixels belonging to the on-ground target. $N$ represents the number of scatterers for each attack. Images with ``Scatterer(s)" are $I(\bm{\Theta}_N,\bm{\xi})$, i.e., the SAR images for the scatterers $\bm{\Theta}_N$. Images with ``Perturbed" are $\mathbf{X}+I(\bm{\Theta}_N,\bm{\xi})$, i.e., the perturbed SAR images after adversarial attacks. The figures show that the scatterer(s) generated from our OTSA satisfy the positioning constraint while the scatterer(s) from the baseline are not on the target.}
  \label{fig:sample}
  \vspace{-0.2cm}
\end{figure*}

\subsection{On-Target Scatterer Adversarial Attack (OTSA)}

Considering the positioning constraint, we propose our scatter-based physical adversarial attack, named \textbf{On-Target Scatterer Adversarial Attack (OTSA)}. For each scatterer parameterized by $\bm{\theta}=[A, x, y, \gamma, L, \alpha, \phi]$, we define a positioning score $S(x,y)$ to encourage it to stay on the target. Let $\mathcal{M}_{\mathbf{X}}=\{(x'_1,y'_1),(x'_2,y'_2),...\}$ be the set of coordinates of all pixels that belong to the target. For a SAR image $\mathbf{X}$, $\mathcal{M}_{\mathbf{X}}$ can be generated by a SAR image segmentation algorithm~\cite{sarbake}.

The score function should be defined to satisfy three requirements: (1) The function must be differentiable in order to allow the optimization problem to be solved by gradient descent algorithms. (2) Pixels on the target should have distinguishably higher scores than those off the target.
(3) All pixels within the target should have the identical score because no position in the target has higher priority than others.

To satisfy the three requirements, we define $S(x,y)$ as:
\begin{equation}
\label{eq:s}
    \Bar{S}(x,y)=\sum_{(x'_j,y'_j)\in \mathcal{M}_{\mathbf{X}}}{\exp{\left(-\frac{1}{2\sigma^2}\lVert(x,y)-(x'_j,y'_j)\rVert_2^2\right)}}
\end{equation}
\begin{equation}
\label{eq:min}
    S(x,y)=\min(\Bar{S}(x,y), \text{MAX})
\end{equation}
where $\lVert\cdot\rVert_2$ is the $L_2$-norm that measures the Euclidean distance between the scatterer's position $(x,y)$ and each pixel $(x'_j,y'_j)$ in $\mathcal{M}$. The score is based on Gaussian kernels that give a high score when the scatterer is close to the on-ground target. The hyperparameter $\sigma$ can be configured to adjust the steadiness of the Gaussian kernels.
To satisfy the third requirement, we apply a $\min(\cdot,\cdot)$ function to truncate the score above a predefined threshold $\text{MAX}$ as shown in Equation~\ref{eq:min}.

Having the positioning score, we define the OTSA as solving an optimization problem. For a given SAR image $\mathbf{X}\in\mathbb{R}^{m\times n}$, to generate a perturbed image $\mathbf{X}^\text{adv}$ with $N$ scatterers, OTSA solves
\begin{align}
\label{eq:opt}
\vspace{-0.2cm}
    \begin{split}
        \arg\max_{\bm{\Theta}_N}\quad &\mathcal{L}(F(\mathbf{X}^\text{adv}), y)+\frac{\lambda}{N}\sum_{i=1}^NS(x_i,y_i) \\
        \text{s.t.}         \quad&\mathbf{X}^\text{adv}=\mathbf{X}+I(\bm{\Theta}_N,\bm{\xi})\in\mathbb{R}^{m\times n}\\
        \quad&\bm{\Theta}_N=\{\bm{\theta}_1,...,\bm{\theta}_N\}\\
        \quad&\bm{\theta}_\text{min}\le\bm{\theta}_i\le\bm{\theta}_\text{max},i=1,...,N    
    \end{split}
\end{align}
where $\mathcal{L}(F(\mathbf{X}^{\text{adv}}),y)$ is the cross-entropy loss measuring the distance between the prediction of the classifier $F(\mathbf{X}^{\text{adv}})$ and the ground truth $y$. The second term is the average positioning score of $N$ scatterers. A constant $\lambda$ balances the importance of the two terms. The $I(\bm{\Theta}_N,\bm{\xi})$ is the SAR image of $N$ scatterers estimated by the ASCM described in Section~\ref{sec:scattering}. $\bm{\Theta}_N=\{\bm{\theta}_1,...,\bm{\theta}_N\}$ are parameters describing $N$ scatterers, and $\bm{\xi}$ is a set of parameters about the SAR imaging, which are summarized in Table~\ref{tab:param}. $\bm{\theta}_\text{min}$ and $\bm{\theta}_\text{max}$ are the minimum and maximum values of each scatterer's parameters in $\bm{\Theta}_N$. 
By solving this optimization problem, our OTSA generates concrete values of $\bm{\Theta}_N$ for up to $N$ scatterers, which instructs the adversary how to physically place the scatterers to implement the attack.
To solve this optimization problem, we use the gradient ascent algorithm. We randomly initialize $N$ scatterers on the target and iteratively update their parameters $\bm{\Theta}_N$ simutaneously. The gradient ascent stops whenever the scatterers are all on the target and the classifier's confidence in the ground truth class is reduced to a low level (e.g., $<10\%$).

\textbf{Remark: }We give an intuitive explanation on why the positioning score works. For each scatterer $\theta_i$, its positioning score $S(x_i,y_i)$ equals $\text{MAX}$ if and only if it is on the target. 
This is true when $\sigma$ is sufficiently small such that the Gaussian kernel in Equation~\ref{eq:s} is very steep and thus any $(x_i,y_i) \notin\mathcal{M}_{\mathbf{X}}$ will have a low score $S(x_i,y_i)$.
Initially, for a scatterer on the target, its score has no gradient with respect to $\theta_i$, i.e., $\nabla_\theta S(x_i,y_i)=0$ due to the $\min$ function in Equation~\ref{eq:min}.
Consequently, the optimization procedure seeks to maximize only the cross-entropy loss, which may inadvertently 
move the scatterer away from the target.
However, when a scatterer is not on the target, its positioning score is less than $\text{MAX}$, and its gradient $\nabla_\theta S(x_i,y_i)$ becomes non-zero.
This non-zero gradient acts to pull the scatterer towards the target, increasing the likelihood that the optimization algorithm will place the scatterer on the target. In summary, the positioning score works by guiding the placement of scatterers towards the target region, resulting in a solution that satisfies the positioning constraint.

\section{Experiments}
\label{sec:experiments}

\subsection{Experimental Setup}

Instead of physically executing attacks, we evaluate our method by simulating the scatterers on the SAR images. This follows the same approach as our baseline \cite{scatter-attack}, a scatterer-based attack without a positioning score (described in Section~\ref{sec:baseline}).
We implement our method using PyTorch 1.11.0 and evaluate it over the Moving and Stationary Target Acquisition and Recognition (MSTAR)~\cite{mstar} dataset. The dataset consists of SAR images of 10 categories of on-ground vehicles, comprising 2747 images for training and 2425 images for testing. The images are one-channel grayscale with 128$\times$128 pixels. 

We perform experiments on nine SAR image classifiers. Of these classifiers, eight are Convolutional Neural Networks (CNNs) and one is a Graph Neural Network (GNN)~\cite{gnn}. 
We follow the standard data preprocessing procedure described in~\cite{preprocessing}, which involves randomly selecting patches of 88$\times$88 pixels from each training image for data augmentation, and using the central patch of 88$\times$88 pixels from each test image for evaluation.
The nine classifiers are trained and evaluated and the accuracies are presented in Table~\ref{tab:classifiers}.

For each SAR image $\bm{x}$, we get $\mathcal{M}_{\bm{x}}$ from the SARBake~\cite{sarbake-dataset} dataset, an overlay on top of MSTAR which marks whether each pixel belongs to the on-ground targets, shallow regions, or irrelevant backgrounds. The SARBake dataset is generated by the SAR image segmentation algorithm proposed in~\cite{sarbake} and thus can be generalized to SAR images other than MSTAR. 

The metric to evaluate the performance of adversarial attacks is the \textbf{success rate} defined as \textit{the ratio of the number of misclassifications to the total number of attacked SAR images}:
\begin{equation}
    \frac{1}{|\mathcal{X}|}\sum_{\mathbf{X}\in \mathcal{X}} \mathbb{I}[F(\mathbf{X}^\text{adv})\ne y_{\mathbf{X}}]
\end{equation}
where $\mathcal{X}$ is the set of input SAR images to manipulate and $y_{\mathbf{X}}$ is the ground truth class of an input SAR image $\mathbf{X}$. The indicator function $\mathbb{I}[\cdot]$ is 1 if the condition inside the brackets is true, and otherwise 0. Note that to make sure all the misclassifications are caused by the attacks rather than the errors of the classifiers themselves, $\mathcal{X}$ excludes input SAR images that are already misclassified before the adversarial attacks.
To generate $\mathcal{X}$, we remove any test images from MSTAR that are misclassified by any of the nine classifiers before performing the attacks. For each of the 10 classes, we randomly sample 10 images to create a set of 100 images to compose $\mathcal{X}$ such that $|\mathcal{X}|=100$.

\subsection{Comparison with the Baseline}

To illustrate the effectiveness of our method, we simulate our OTSA attack on the 100 sampled images using the nine classifiers respectively.
We let $\bm{\theta}_\text{min}=[0,0,0,-1,0,0,-1]$ and $\bm{\theta}_\text{max}=[10,87,87,1,2,5,1]$ following the values in~\cite{scatter-attack}. We also set $\sigma=0.4$, $\text{MAX}=0.5$ and $\lambda=10$ as they lead to the best performance in our experiments.

Figure~\ref{fig:comparison}, with red and purple bars, shows the experimental results of our OTSA attack and the baseline attack using 1, 2, and 3 scatterers, respectively. 
We compare success rates from the two methods while 
enforcing the scatterer positioning constraint,
i.e., 
before evaluating the outcome of the attacked image, we filter out the scatterers that are not on the target. We observe significant improvements in the success rate of our attacks compared with the baseline when the positioning constraint is enforced.

In Figure~\ref{fig:sample}, we show an example input SAR image and its perturbations generated by our OTSA attack and the baseline. We also show a bitmap image representing the pixels belonging to the target region where the scatterers should be constrained in. As shown in the figure, the scatterers generated by the baseline method fall into the shadow regions, while those generated by our OTSA attack are on the target.

\section{Related Works}

\subsection{Traditional Adversarial Attacks}

The general idea is to increase the loss of a classifier $\mathcal{L}(F(\mathbf{X}^{\text{adv}}),y)$
that measures the distance between the prediction of the classifier $F(\mathbf{X}^{\text{adv}})$ and the ground truth $y$. To achieve this, a typical algorithm is the fast gradient sign method (FGSM)~\cite{fgsm} which generates a perturbed image as 
$\mathbf{X}^\text{adv}=\mathbf{X}+\epsilon\cdot\text{sgn}(\nabla_{\mathbf{X}}\mathcal{L}(F(\mathbf{X}),y))$,
where $\text{sgn}(\cdot)$ is the sign function, and $\epsilon$ is a small constant to make the perturbation imperceptible to human eyes.
Based on FGSM, a sequence of variants are also developed, such as iterative FGSM~\cite{i-fgsm}. 
Another class of approaches formulate the attack as an optimization problem, e.g., minimizing
$-\mathcal{L}(F(\mathbf{X}^\text{adv}),y)+\lambda\lVert\mathbf{X}^\text{adv}-\mathbf{X}\rVert_p$,
which increases the loss while constraining the the perturbation
by $L_p$-norm 
with a factor $\lambda$ balancing the relative importance of the two terms. 
As mentioned in Section~\ref{sec:threat}, these methods for natural images are not feasible for SAR ATR. In constrast, our OTSA is a  physical adversarial attack with much better feasibility.

\subsection{Adversarial Attack for SAR Image Classifiers}
\label{sec:baseline}

Under the same threat model we defined in~\ref{sec:threat}, a scatterer-based adversarial attack~\cite{scatter-attack} was proposed. 
Like our method, the ASCM described in Section~\ref{sec:scattering} is utilized to estimate the SAR image for the scatterers. Their attack is to solve the following optimization problem
\begin{align}
    \label{eq:old-opt}
    \begin{split}
        \arg\max_{\bm{\Theta}_N}\quad &\mathcal{L}(F(\mathbf{X}^\text{adv}), y) \\
        \text{s.t.} 
        \quad&\mathbf{X}^\text{adv}=\mathbf{X}+I(\bm{\Theta}_N,\bm{\xi})\in\mathbb{R}^{m\times n}\\
        \quad&\bm{\Theta}_N=\{\bm{\theta}_1,...,\bm{\theta}_N\}\\
        \quad&\bm{\theta}_\text{min}\le\bm{\theta}_i\le\bm{\theta}_\text{max},i=1,...,N              
    \end{split}
\end{align}
which is similar to our OTSA except the missing of the positioning score.
Different from our OTSA, this method does not consider the positioning constraint. In Section~\ref{sec:motivation}, we have discussed the
problems of not constraining the scatterer positions and the benefits of enforcing the positioning constraint. In Section~\ref{sec:experiments}, we showed that the success rate of this method is lower than our OTSA.

\section{Conclusion}
In this work, we revisited the physical adversarial attack problem on SAR image classifiers and proposed the On-Target Scatterer Attack (OTSA). Our approach enables the attacker to place false objects as scatterers near the on-ground targets to perturb the SAR images. The constraint on the positioning of the scatterers improves the feasibility of this approach. To enforce the constraint, we proposed a positioning score function based on Gaussian kernels and formulate OTSA as solving an optimization problem. We conducted experiments to attack nine different classifiers over the MSTAR dataset with our OTSA attack. We obtained significant improvements in terms of the success rate of attacks compared with the baseline. This work is a cornerstone of future works on adversarial attacks on SAR images for moving on-ground targets. In the future, we will also integrate a 3-D scattering model rather than the 2-D ASCM into attacks to more precisely simulate how 3-D scatterers appear in SAR images.

\section*{Acknowledgement}
This work is supported by the DEVCOM Army Research Lab (ARL) under grant W911NF2220159.

\bigskip

\textbf{Distribution Statement A:} Approved for public release. Distribution is unlimited.

\bibliographystyle{IEEEtran}
\bibliography{main}

\end{document}